\documentclass[runningheads]{llncs}
\usepackage{cite}

\usepackage[pdftex]{graphicx}
\graphicspath{{figures/}}
\DeclareGraphicsExtensions{.pdf,.jpeg,.png}

\usepackage[cmex10]{amsmath}
\usepackage{xfrac}
\usepackage{amssymb}
\usepackage{amsfonts}
\usepackage{accents}
\DeclareMathAlphabet{\mymathbb}{U}{bbold}{m}{n}

\usepackage{algorithmic}

\usepackage{array}
\usepackage{multirow}

\usepackage{enumitem}

\usepackage{url}
\usepackage{hyperref}

\usepackage{xcolor}

\begin{document}
\title{Approximation of dilation-based spatial relations to add structural constraints in neural networks\thanks{Supported by Agence Nationale de la Recherche (ANR), project ANR-17-CE23-0021, and São Paulo Research Foundation (FAPESP), projects 2017/50236-1 and 2015/22308-2. F. Yger acknowledges the support of the ANR as part of the ``Investissements d'avenir" program, reference ANR-19-P3IA-0001 (PRAIRIE 3IA Institute).}}
%
%
\author{Mateus Riva\inst{1} \and
Pietro Gori\inst{1} \and
Florian Yger\inst{2}\orcidID{0000-0002-7182-8062}\and Roberto Cesar\inst{3}\orcidID{0000-0003-2701-4288} \and
Isabelle Bloch\orcidID{0000-0002-6984-1532}\thanks{This work was partly done while I. Bloch was with LTCI, T\'el\'ecom Paris, Institut Polytechnique de Paris, France.}}
\titlerunning{Approximation of dilation-based spatial relations}
\authorrunning{M. Riva et al.}
%
\institute{
    LTCI, Télécom Paris, Institut Polytechnique de Paris, France\\
    \email{\{mateus.riva, pietro.gori\}@telecom-paris.fr} \and
    LAMSADE, Université Paris-Dauphine, PSL Research University,  France\\
    \email{florian.yger@dauphine.fr} \and
    IME, University of São Paulo, Brazil\\
    \email{rmcesar@usp.br} \and
    Sorbonne Université, CNRS, LIP6, F-75005 Paris, France \\
\email{isabelle.bloch@sorbonne-universite.fr}
}
\maketitle              
\begin{abstract}
Spatial relations between objects in an image have proved useful for structural object recognition. Structural constraints can act as regularization in neural network training, improving generalization capability with small datasets. Several relations can be modeled as a morphological dilation of a reference object with a structuring element representing the semantics of the relation, from which the degree of satisfaction of the relation between another object and the reference object can be derived.
However, dilation is not differentiable, requiring an approximation to be used in the context of gradient-descent training of a network. We propose to approximate dilations using convolutions based on a kernel equal to the structuring element. We show that the proposed approximation, even if slightly less accurate than previous approximations, is definitely faster to compute and therefore more suitable for computationally intensive neural network applications.
\end{abstract}
\section{Introduction}
Recently, the computer vision and image processing literature has met a veritable deluge of papers applying and exploring neural network-based techniques to a wide array of fields and problems. While these techniques often show remarkable accuracy and return high-quality results, they invariably require large amounts of data to train the vast array of parameters of a typical neural network. In 
domains where the amount of annotated data is limited, unless mitigated by other techniques, this ``data-hunger'' may result in sub-optimal neural networks performance.
Regularization is a useful tool for allowing greater generalization from learning on limited datasets. In particular, regularization can be performed by introducing domain-specific constraints based on reliable expectations on the nature of the data, such as structural constraints.

In many domains in computer vision, such as e.g. medical imaging, the task of semantic segmentation of objects in a scene may be guided by the use of  structural constraints based on prior knowledge (e.g. on the spatial organization of the objects). Imposing such constraints as a form of regularization may help a machine learning system in leveraging the structural nature of the images for improving segmentation and analysis quality, while requiring a smaller quantity of data. For doing so, it is necessary to have both a structural model that encodes the prior information, and a method for usefully embedding this structural model into the machine learning process.

Many useful structural constraint models are based on mathematical morphology operators (in particular, dilation), which are however non-differentiable over the entire domain and thus unable to be properly used in the context of gradient-descent learning -- the backbone of most neural network training methods. With the intention of leveraging these useful models for the improvement of neural network training, we propose a comparative study of convolutional-based approximations of the mathematical morphology dilation, as the first step towards realizing this ambition.

In this work we present a comparative study of differentiable approximations of the dilation operator for spatial relationship encoding in the context of neural network training.
We propose a simple, non-parametrized approximation through the use of the convolution, and compare it with known approximations such as mean operators. 

%
%
\section{Related Work}
\subsection{Fuzzy mathematical morphology for structural modeling}
The main inspiration for this work was the review on fuzzy spatial relationships in~\cite{bloch_fuzzy_2005}. The representative power of fuzzy spatial relationships has great potential for application to modern deep learning spatial awareness techniques, such as multi-task spatial relationship learning~\cite{murugesan_psi-net:_2019} or spatial-based loss regularization addition~\cite{Simantiris2020-STSP}.
The fuzzy sets theory is an appropriate framework for information representation and processing, taking into account its intrinsic imprecision. In the present context for structural scene understanding, imprecision can pertain to both objects, which are then represented as fuzzy sets in the spatial domain $\Omega$, and to spatial relations, which will then hold to some degree.
Since the output of a typical semantic segmentation neural network is a set of per-pixel class probabilities, there is a small paradigm shift to see these as an ensemble of fuzzy sets with further constraints. 
Several spatial relations can be modeled using fuzzy mathematical morphology, in particular fuzzy dilation. The main idea is to represent the semantics of a given relation as a fuzzy structuring element $w$ in the spatial domain, and dilate a reference object $m$ by this structuring element in order to define the region of space where the relation to this object is satisfied. The fuzzy dilation is defined as~\cite{IB:FSS-09}:
$\delta_w m(x) = \sup_{y\in w(y)} t(w(x-y), m(y))$, where $t$ is a t-norm, and $\delta_w m(x)$ is the degree to which the relation to $m$ is satisfied at point $x$.
Several relations can be defined according to this general principle. In this work we will focus on closeness and directional relations.



\subsection{Differentiable approximations of mathematical morphology operators}


A differentiable approximation of $\delta_w m$ requires an approximation of the $\max$ function. Following Dubois and Prade~\cite[Sect. 1.2.1]{Dubois1985-IS-aggregation}, the $\max$ can be approximated as the limit case of the generalized mean (for any $x, y \in [0,1]$):
\begin{equation}
    \label{eq:mean_max}
    \max(x,y) = \lim_{p\to\infty} \left( \frac{x^p + y^p}{2} \right)^{\sfrac{1}{p}}
\end{equation}
Thus we can rewrite the dilation operation $\delta_w m$ as:
\begin{equation}
 \label{eq:mean_max_approx}
    \delta_w m(x) = \lim_{p\to\infty} \left( \frac{\int_{y\in \Omega} t(w(x-y),m(y))^p dy}{\int_{y\in \Omega} w(y) dy} \right)^{\sfrac{1}{p}}
\end{equation}
where $t$ is a derivable t-norm (such as the product), and we can approximate it using a (moderately) large positive value of $p$ in Equation~\ref{eq:mean_max_approx}. A value of $p$ as low as $70$ produces errors below $0.01$ in Equation~\ref{eq:mean_max} when $x$ and $y$ are scalars.
Note that in practice, $\Omega$ is a bounded finite domain, and $w$ has a finite support, hence the integral converges.

Another common approximation is to use the limit cases of the counter-harmonic mean (CHM) as approximations of the fundamental morphological erosion and dilation operators~\cite{Angulo2010-ACIVS-morphconv-chm, Masci2013-MMASIP-morphconv-chm}. Additionally, this approximation has been applied already in deep learning contexts~\cite{Mellouli2019-NNLS-morphconv-chm}. The CHM $n$ at the $p$-th power of an image $m$ with structuring element $w$ is defined as:
\begin{equation}
    \label{eq:chm_definition}
    n^p_wm(x) = \frac{(m^{p+1} \ast w)(x)}{(m^p \ast w)(x)} = \frac{\int_{y \in w(x)}m^{p+1}(y) w(x-y)dy}{\int_{y \in w(x)}m^{p}(y) w(x-y)dy}
\end{equation}
where $m^p$ is the image $m$ with every pixel elevated to the power $p$ and $\ast$ denotes the convolution operation.
The morphological dilation  $\delta_w m$, can thus be seen as the limit case of the CHM:
\begin{equation}
    \label{eq:chm_dilation}
    \delta_w m(x) = \lim_{p\to\infty} n^p_wm(x) 
\end{equation}
and can be approximated by taking a positive value of $p$ in Equation~\ref{eq:chm_definition}. A value of $p$ as low as $30$ produces errors below $0.01$ for the counter-harmonic mean of two scalars.


%
%
\section{Methodology}
\begin{figure}
    \centering
    \begin{tabular}{ccc}
        \includegraphics[width=0.18\textwidth]{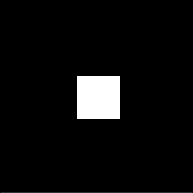} & 
        \includegraphics[width=0.18\textwidth]{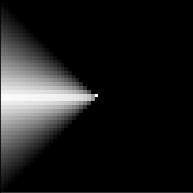} & 
        \includegraphics[width=0.18\textwidth]{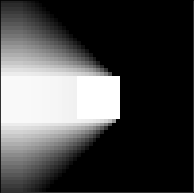}  \\
        Source & Kernel & Relational Map
    \end{tabular}
    \caption{Examples of a source, a structuring element (here encoding ``to the left of'') and the relational map produced by the dilation of the source object by this structuring element.}
    \label{fig:rm_examples}
\end{figure}

As in~\cite{bloch_fuzzy_2005}, we propose to encode expected \textit{a priori} relative positioning as a set of fuzzy mathematical morphology structuring elements -- each element representing the semantics of a relationship. Each relationship is defined between two specific objects, which we call \emph{source} and \emph{target}. The dilation of the source by the structuring element results in a fuzzy ``map'' that is high-valued in regions obeying the expected relative positioning, as can be seen in Figure~\ref{fig:rm_examples}. The intersection of this map with the target object allows us to compute the satisfaction degree of the relationship.

\subsection{Fuzzy Spatial Relationships}
In this paper we focus, as examples, on two interesting relationships: closeness and directional relative position between two objects~\cite{bloch_fuzzy_2005}.

\subsubsection{Closeness.} One of the simple types of relationship between objects is ``closeness'' or proximity, that is, how distant one object is from the other. While there are several measures and metrics that propose to solve this problem, the relational maps offer an interesting solution: the dilation of the source by a circular structuring element of radius $r$ will result in a relational map with positive values up to $r$ from the source, thus encoding the relationship ``at most $r$ pixels from the source''. This extends directly to a non-crisp structuring element, modeling the intrinsic imprecision of the concept ``close to'', where the membership of a point is a decreasing function of the distance to the origin. The degree to which a
target object is ``close to'' the source is then measured as a degree of intersection with this map.
The use of a ring shaped structuring element may also additionally enforce a minimal distance between objects as the inner radius of the structuring element.

\subsubsection{Directional Relative Position.} The directional relative position between objects allows us to encode an expected configuration of a scene, particularly useful in domains where scenes are highly constrained such as anatomical images. Encoding such a relationship can be done via the dilation of the source object with a structuring element defined at point $x$ as a decreasing function of the angle between the segment joining the origin and $x$ and the line in the desired direction (see  Figure~\ref{fig:rm_examples} for the left direction).


%

\subsubsection{Defining Fuzzy Spatial Relationships.} Given a relationship $\rho=(k,l,B_{kl})$, where $k$ is an image containing the source object, $l$ is an image containing the target object, $B_{kl}$ a relationship structuring element (or kernel) which encodes the relationship of the source with a target object and $\Omega$ their domain spaces (for simplifying calculation, we assume all domains are shared\footnote{Please note that this does not impact the generalizability of the proposed algorithm.}), we can build the dilation-based relational map of $\rho$, $\Phi_\delta^{(\rho)}$, by dilating $k$ with $B_{kl}$. The dilation-based intersected relational map $\Psi_\delta^{(\rho)}$ is obtained by the element-wise multiplication -- here denoted as $(\cdot)$ -- of $\Phi_\delta^{(\rho)}$ with $l$. Finally, the satisfaction degree of the relationship is given by the normalized relational score $S_\delta^{(\rho)}$, obtained by the division of the sum of all values in $\Psi_\delta^{(\rho)}$ by the sum of all values in $l$:
\begin{equation}
    \label{eq:phi_delta}
   \forall x \in \Omega, \; \Phi_\delta^{(\rho)}(x) = \delta_{B_{kl}}k(x)
 \quad \mbox{and} \quad  \Psi_\delta^{(\rho)}(x) = \Phi_\delta^{(\rho)}(x) \cdot l(x)
\end{equation}
\begin{equation}
    \label{eq:s_delta}
    S_\delta^{(\rho)} = \frac{\sum_{x\in\Omega}\Psi_\delta^{(\rho)}(x)}{\sum_{x\in\Omega}l(x)}
\end{equation}

\subsection{Differentiable Approximation}
In the context of neural network learning, these maps or scores may be used to compute an additional loss term to guide the training. However, as the loss needs to be differentiable with respect to the objects memberships to be predicted, the morphological dilation must be approximated by a differentiable operator.

For the purpose of encoding the spatial relationship, the dilation can be approximated by a simple convolution, with the kernel being the structuring element flipped across all axes. For clarity, we will refer to both kernels as $B_{kl}$. The convolution is a practical choice in the context of neural networks. Not only convolutions are highly optimized, due to the popularity of convolutional neural networks (CNNs), which allow for faster calculation, but also the memory consumption decreases when compared to  mean approximations (due to the power functions). Additionally, the convolutional approximation is hyperparameter-free.

Given a relationship $\rho=(k,l,B_{kl})$ as defined above, we can build the convolution-approximated relational map of $\rho$, $\Phi_\ast^{(\rho)}$, by convolving $k$ by $B_{kl}$. The approximated intersected relational map $\Psi_\delta^{(\rho)}$ and normalized relational score are then derived as before:
\begin{equation}
    \label{eq:phi_ast}
    \forall x \in \Omega, \; \Phi_\ast^{(\rho)}(x) = (k \ast B_{kl})(x)
\quad \mbox{and} \quad \Psi_\ast^{(\rho)}(x) = \Phi_\ast^{(\rho)} (x) \cdot l(x)
\end{equation}
\begin{equation}
    \label{eq:s_ast}
    S_\ast^{(\rho)} = \frac{\sum_{x\in\Omega}\Psi_\ast^{(\rho)}(x)}{\sum_{x\in\Omega}l(x)}
\end{equation}

Such scores, for several $\rho$, are then included in the loss function, with the aim of maximizing them.
This optimization is done by differentiating the loss function with respect to every object membership function. 
This proposed approximation is indeed differentiable, as a composition of differentiable functions. This allows its use in gradient descent-based learning algorithms.
Obviously, only derivatives of $S_\ast^{(\rho)}$ with respect to $k$ and $l$ will be non-zero if the relation involves only objects $k$ and $l$.
Let us now detail the derivative calculations.
The derivative of the approximated normalized relational score with respect to $m(x)$, for an object $m$, is:
\begin{multline}
\label{eq:s_derivative}
    \frac{\partial}{\partial m(x)} S_\ast^{(\rho)} = \frac{\partial}{\partial m(x)}  \frac{\sum_{u\in\Omega}\Psi_\ast^{(\rho)}(u)}{\sum_{u\in\Omega}l(u)} = \\
    \frac{\sum_{u\in\Omega} \frac{\partial}{\partial m(x)}(\Psi_\ast^{(\rho)}(u))\sum_{u\in\Omega}l(u) - \sum_{u\in\Omega}\frac{\partial}{\partial m(x)}(l(u)) \sum_{u\in\Omega}\Psi_\ast^{(\rho)}(u)}{\left(\sum_{u\in\Omega}l(u)\right)^2}
\end{multline}
The derivative of the approximated intersected relation map
is:
\begin{equation}
\label{eq:psi_derivative}
    \frac{\partial}{\partial m(x)} \Psi_\ast^{(\rho)}(u) = \Phi_\ast^{(\rho)}(u)\frac{\partial}{\partial m(x)} l(u) + l(u)\frac{\partial}{\partial m(x)}(k\ast B_{kl})(u)
\end{equation}


\paragraph{Case $m \neq k, m \neq l$.} As mentioned above, the derivative is equal to 0.

\paragraph{Case $m = l$.} By noting that $\frac{\partial}{\partial l(x)} \Phi_\ast^{(\rho)}(u)=0, \forall u\in \Omega$,
Equation~\ref{eq:psi_derivative} becomes:
\begin{equation}
    \label{eq:psi_derivative_l}
    \frac{\partial}{\partial l(x)} \Psi_\ast^{(\rho)}(u) = \left\{
    \begin{array}{ll}
    \Phi_\ast^{(\rho)}(u) & \mbox{if } u=x\\
    0& \mbox{otherwise}
    \end{array}\right.
\end{equation}
%
and we get:
\begin{equation}
\frac{\partial}{\partial l(x)} S_\ast^{(\rho)} =\frac{ \Phi_\ast^{(\rho)}(x) \sum_{u\in\Omega}l(u) - \sum_{u\in\Omega}(\Psi_\ast^{(\rho)}(u))}{\left(\sum_{u\in\Omega}l(u)\right)^2}
\end{equation}

\paragraph{Case $m = k$.} The first term in Equation~\ref{eq:psi_derivative} is equal to 0, and we have:
\begin{equation}
\label{eq:phi_derivative_k}
\frac{\partial}{\partial k(x)}(k\ast B_{kl})(u) =
    \frac{\partial}{\partial k(x)}(\sum_{t \in \Omega} k(t)B_{kl}(u-t)) = B_{kl}(u-x)
\end{equation}
Equation~\ref{eq:psi_derivative} then becomes
$    \frac{\partial}{\partial k(x)} \Psi_\ast^{(\rho)}(u) = l(u)B_{kl}(u-x)$,
and finally we get:
\begin{equation}
    \label{eq:s_derivative_k}
    \frac{\partial}{\partial k(x)} S_\ast^{(\rho)} =    \frac{\sum_{u\in\Omega} l(u) B_{kl}(u-x)\sum_{u\in\Omega}l(u) }{\left(\sum_{u\in\Omega}l(u)\right)^2} = 
     \frac{\sum_{u\in\Omega} l(u) B_{kl}(u-x)}{\sum_{u\in\Omega}l(u)}
\end{equation}
%
%
\section{Experiments}
To illustrate that the proposed approximation behaves similarly to the original dilation-based spatial relation encoding,
%
in the context of machine learning, and particularly as a component in a loss function,
we compare the behavior of the relational score obtained by the dilation-based approach, $S_\delta = \sum_{x\in \Omega}(\delta_{B_{kl}}(k)(x))l(x) / \sum_{x\in \Omega}l(x)$, to the score obtained by the convolutional approximation, $S_\ast = \sum_{x\in \Omega}((B_{kl} * k)(x))l(x) / \sum_{x\in \Omega}l(x)$, and in an analogous fashion for the score obtained by the CHM approximation, $S_{CHM}$, and by the generalized mean approximation, $S_\mu$.

\subsection{Comparison -- Experimental Setup}
Given a previously specified source $k$ and relationship kernel $B_{kl}$, we compute the relational maps $\Phi_\delta$ (dilation-based), $\Phi_\ast$ (convolution approximation), $\Phi_{CHM}$ (CHM approximation) and $\Phi_\mu$ (generalized mean approximation). Subsequently, the relational scores $S_\delta$, $S_\ast$, $S_{CHM}$ and $S_\mu$ will be computed for a target $l$ placed on all possible positions $x \in \Omega$. Our expectation is that all functions will behave similarly in similar regions of $\Omega$, with the exception of the source region, which the dilations have a stronger tendency to include.
We use the relations ``to the right of'' a disk of radius 5 pixels placed at the left of the central horizontal line, at coordinates $(20,50)$, closeness with respect to a disk of radius 5 pixels placed at the image center, with a crown-shaped kernel, and an additional ``insidness'' relation to a square of side 50 pixels, using a simple dot kernel.



For all experiments, the sources and targets were placed inside a $100 \times 100$ image, $p$ values were $100$ for CHM and generalized mean. We obtained the dilation-based and approximations-based relational scores for a target at all positions $x \in \Omega$.

\subsubsection{Experimental Results.}
\begin{figure}
    \centering
    \begin{tabular}{|c|c|c|c|c|c|c|}
        \hline
         & Source & Kernel & $\Phi_\delta$ & $\Phi_\ast$ & $\Phi_{CHM}$ & $\Phi_\mu$ \\
        \hline
        Right & \includegraphics[width=0.135\textwidth]{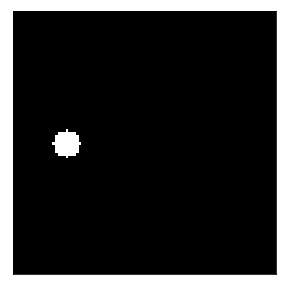} & 
        \includegraphics[width=0.135\textwidth]{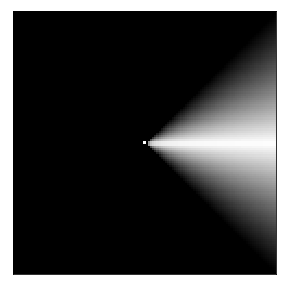} & 
        \includegraphics[width=0.135\textwidth]{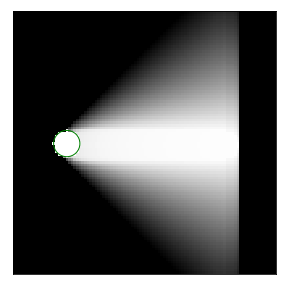} & 
        \includegraphics[width=0.135\textwidth]{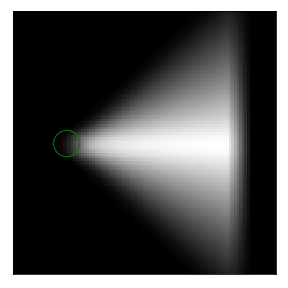} & 
        \includegraphics[width=0.135\textwidth]{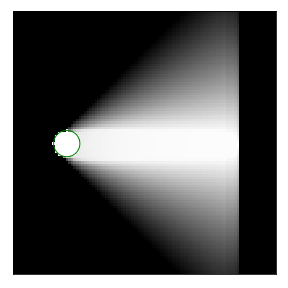} & 
        \includegraphics[width=0.135\textwidth]{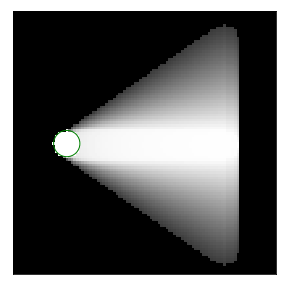}\\
        \cline{2-7}
        & \multicolumn{3}{r|}{Difference to $\Phi_\delta$: } & \includegraphics[width=0.135\textwidth]{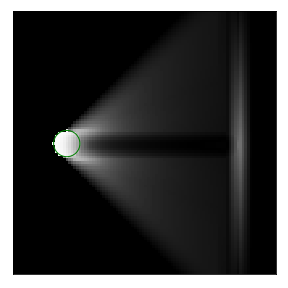} &
        \includegraphics[width=0.135\textwidth]{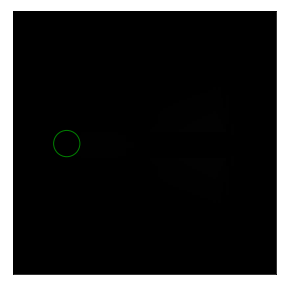}& 
        \includegraphics[width=0.135\textwidth]{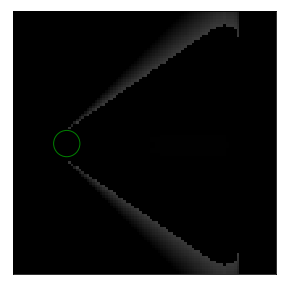}\\
         \hline
        Close to & \includegraphics[width=0.135\textwidth]{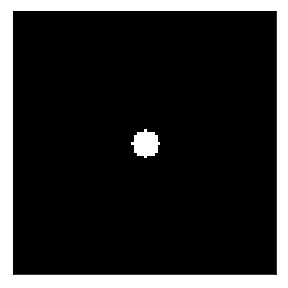} & 
        \includegraphics[width=0.135\textwidth]{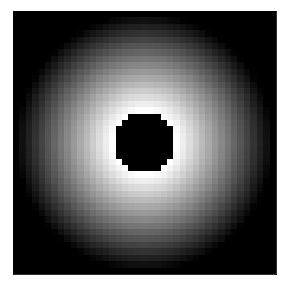} & 
        \includegraphics[width=0.135\textwidth]{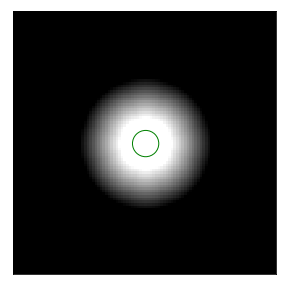} & 
        \includegraphics[width=0.135\textwidth]{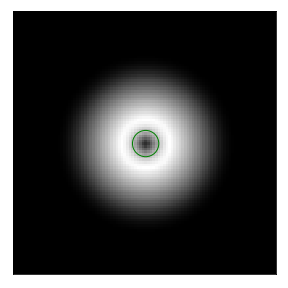} & 
        \includegraphics[width=0.135\textwidth]{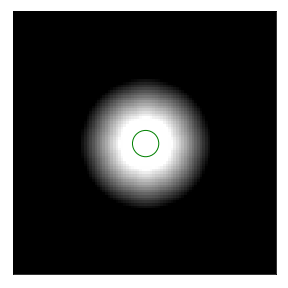} & 
        \includegraphics[width=0.135\textwidth]{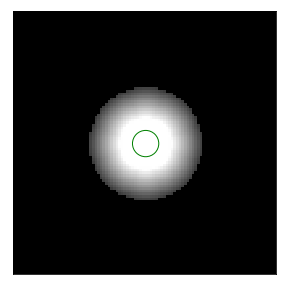}\\
        \cline{2-7}
        & \multicolumn{3}{r|}{Difference to $\Phi_\delta$: } & \includegraphics[width=0.135\textwidth]{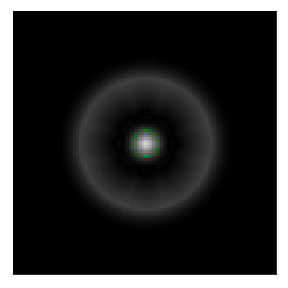} &
        \includegraphics[width=0.135\textwidth]{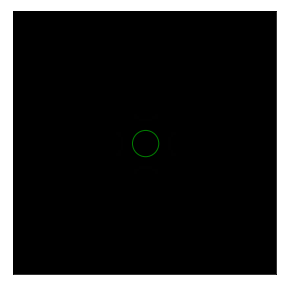} & 
        \includegraphics[width=0.135\textwidth]{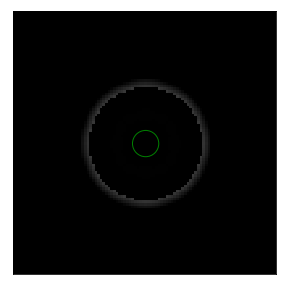}\\
        \hline
        Far from & \includegraphics[width=0.135\textwidth]{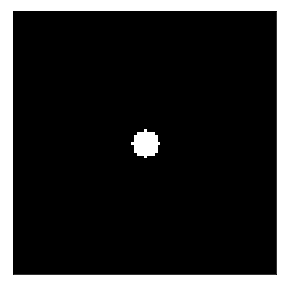} & 
        \includegraphics[width=0.135\textwidth]{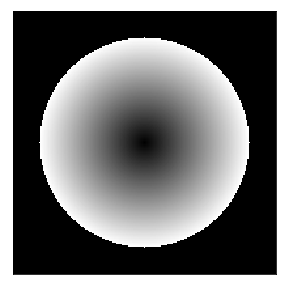} & 
        \includegraphics[width=0.135\textwidth]{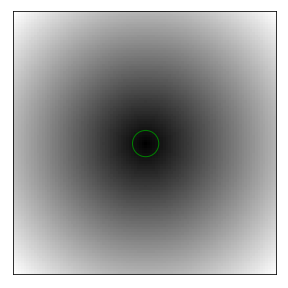} & 
        \includegraphics[width=0.135\textwidth]{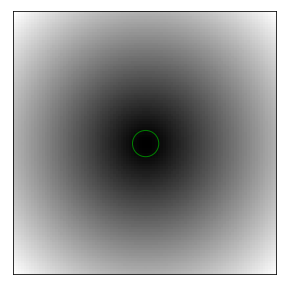} & 
        \includegraphics[width=0.135\textwidth]{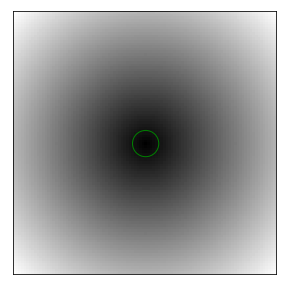} & 
        \includegraphics[width=0.135\textwidth]{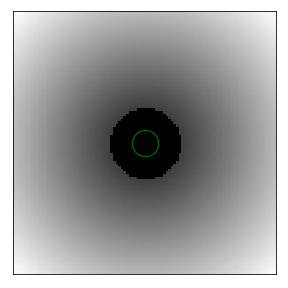}\\
        \cline{2-7}
        & \multicolumn{3}{r|}{Difference to $\Phi_\delta$: } & \includegraphics[width=0.135\textwidth]{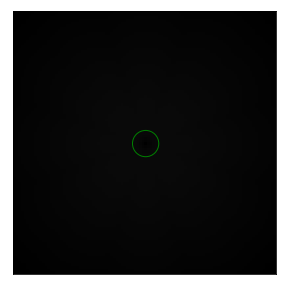} &
        \includegraphics[width=0.135\textwidth]{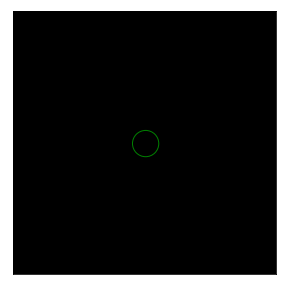}& 
        \includegraphics[width=0.135\textwidth]{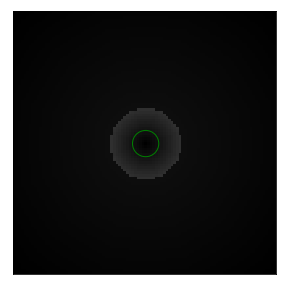}\\
        \hline
        Inside of & \includegraphics[width=0.135\textwidth]{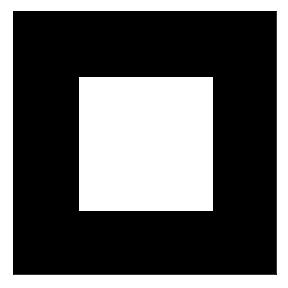} & 
        \includegraphics[width=0.135\textwidth]{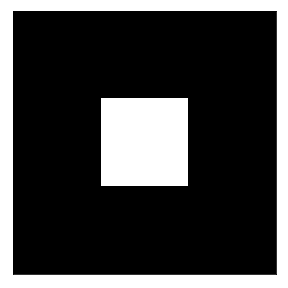} & 
        \includegraphics[width=0.135\textwidth]{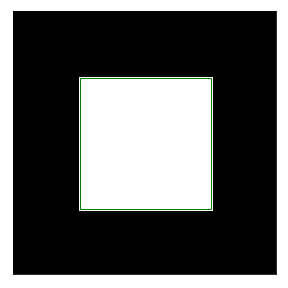} & 
        \includegraphics[width=0.135\textwidth]{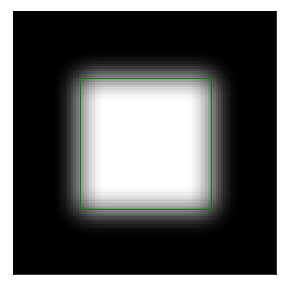} & 
        \includegraphics[width=0.135\textwidth]{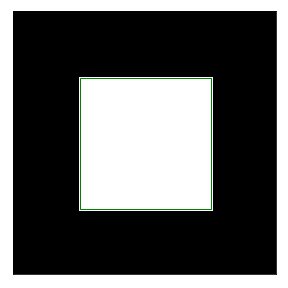} & 
        \includegraphics[width=0.135\textwidth]{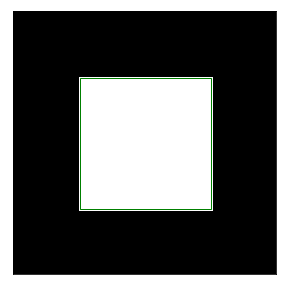}\\
        \cline{2-7}
        & \multicolumn{3}{r|}{Difference to $\Phi_\delta$: } & \includegraphics[width=0.135\textwidth]{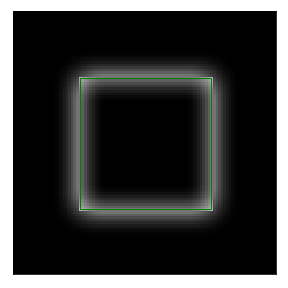}&
        \includegraphics[width=0.135\textwidth]{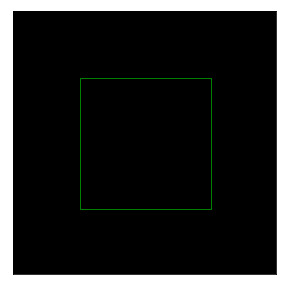}& 
        \includegraphics[width=0.135\textwidth]{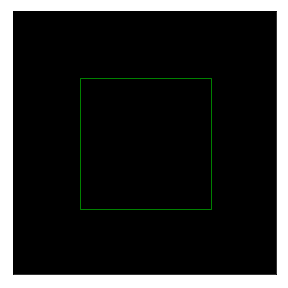}\\
        \hline
    \end{tabular}
    \caption{Comparison of the relational maps obtained using different techniques. See text for notations.
    The green outlines represent the source position in the relational map. The absolute difference to the dilation-based map can be seen for each method. Kernels are rescaled for visualization purpose.}
    \label{fig:phi_comparison}
\end{figure}

\begin{figure}[htbp]
    \centering
    \begin{tabular}{|c|c|c|c|c|}
        \hline
         & $S_\delta$ & $S_\ast - S_\delta$ & $S_{CHM}-S_\delta$ & $S_\mu-S_\delta$ \\
        \hline
        Right & \includegraphics[width=0.135\textwidth]{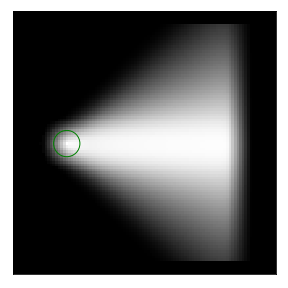} & \includegraphics[width=0.135\textwidth]{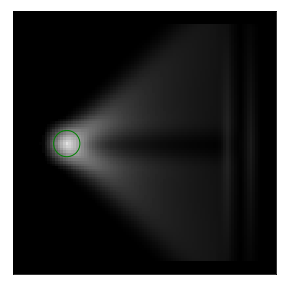}&
        \includegraphics[width=0.135\textwidth]{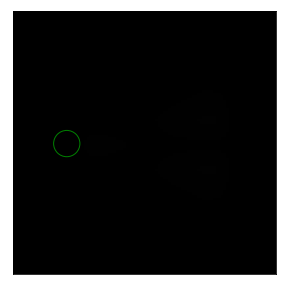}& 
        \includegraphics[width=0.135\textwidth]{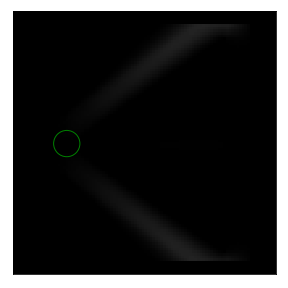}\\
        \hline
        Close to & \includegraphics[width=0.135\textwidth]{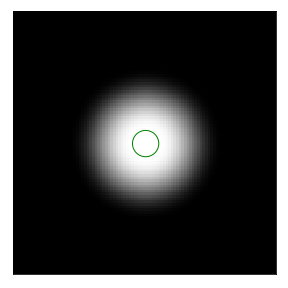} &  \includegraphics[width=0.135\textwidth]{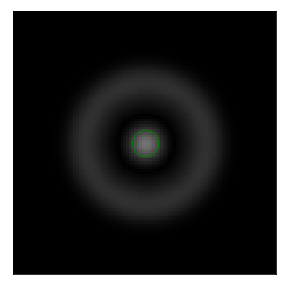}&
        \includegraphics[width=0.135\textwidth]{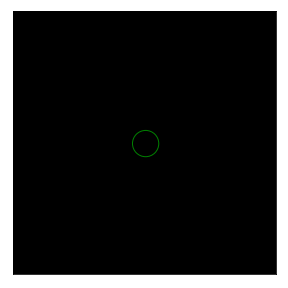}& 
        \includegraphics[width=0.135\textwidth]{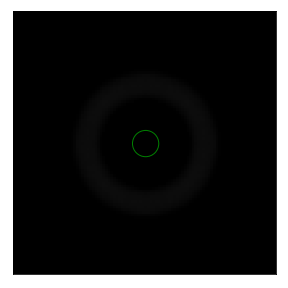}\\
        \hline
        Far from & \includegraphics[width=0.135\textwidth]{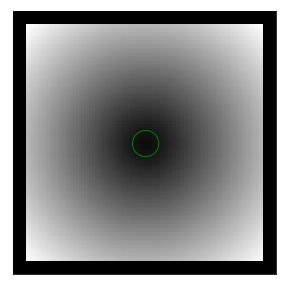} &  \includegraphics[width=0.135\textwidth]{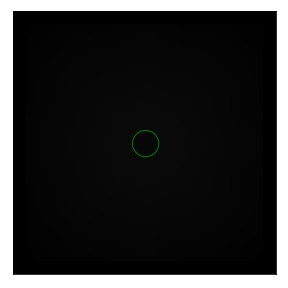}&
        \includegraphics[width=0.135\textwidth]{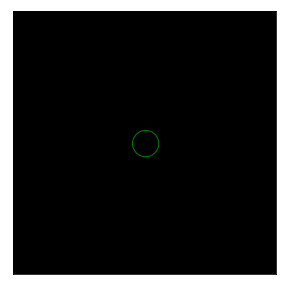}& 
        \includegraphics[width=0.135\textwidth]{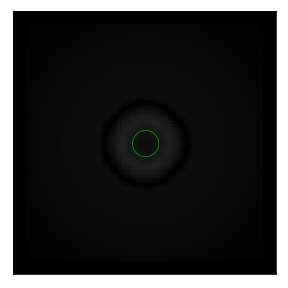}\\
        \hline
        Inside of & \includegraphics[width=0.135\textwidth]{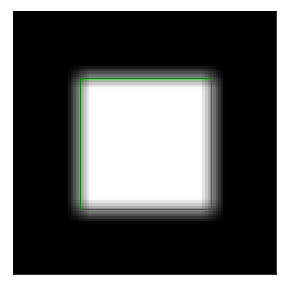} & \includegraphics[width=0.135\textwidth]{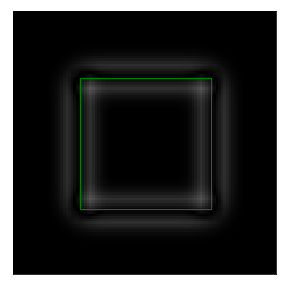}&
        \includegraphics[width=0.135\textwidth]{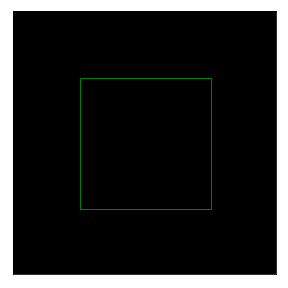}& 
        \includegraphics[width=0.135\textwidth]{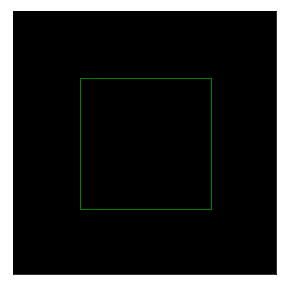}\\
        \hline
    \end{tabular}
    \caption{Comparison of the difference of the relational scores obtained using different techniques. 
    The green outlines represent the source position in the relational map.}
    \label{fig:s_comparison}
\end{figure}

\begin{figure}[htbp]
    \centering
    \begin{tabular}{|c|c|c|}
    \hline
         & Mid-X cut & Mid-Y cut \\
        \hline
        Right & \includegraphics[width=0.3\textwidth]{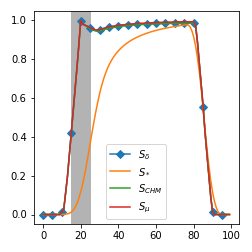}
        & \includegraphics[width=0.3\textwidth]{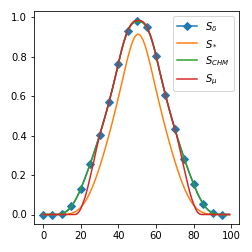}\\
        \hline
        Close to & \includegraphics[width=0.3\textwidth]{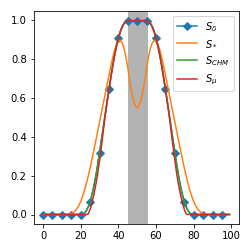}
        & \includegraphics[width=0.3\textwidth]{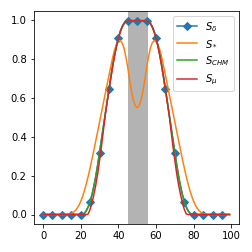}\\
        \hline
        Far from & \includegraphics[width=0.3\textwidth]{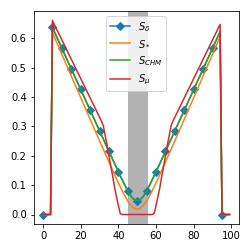}
        & \includegraphics[width=0.3\textwidth]{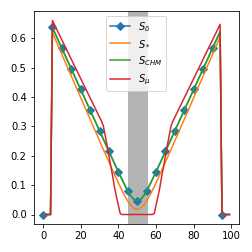}\\
        \hline
        Inside of & \includegraphics[width=0.3\textwidth]{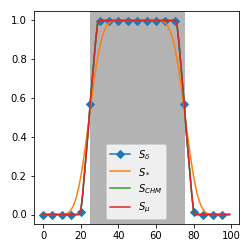}
        & \includegraphics[width=0.3\textwidth]{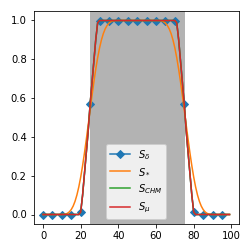}\\
        \hline
    \end{tabular}
    \caption{Comparison of the relational scores obtained using different techniques. The first column is the value of the relational score heatmap at the mid-X-axis cut for all techniques; the second column is for the mid-Y-axis cut. The gray zone represents overlap with the source region. $S_\delta$ and $S_{CHM}$ are visually indistinguishable in all plots.}
    \label{fig:s_midcuts}
\end{figure}

For each experiment conducted, we display, for all techniques, the relational maps (in Figure~\ref{fig:phi_comparison}) and the heatmap of relational score values per target position (in Figure~\ref{fig:s_comparison}). Additionally, we directly compare the curves of all relational score functions at the cut obtained at the center of the X-axis and of the Y-axis (in Figure~\ref{fig:s_midcuts}).

\paragraph{Relative position.} Results for the relative position experiment are shown in the first row of Figures~\ref{fig:phi_comparison}, \ref{fig:s_comparison} and \ref{fig:s_midcuts}. From both the heatmaps and the curves shown, it can be seen that the convolutional approximation results in a slight ``feather'' effect being applied to the farthest points of the valid target region. Additionally, on almost all points $S_\ast$ is attenuated w.r.t. $S_\delta$ but displays the same behavior. The CHM approaches the dilation almost perfectly, with the generalized mean approximation following behind.
Note that the region inside the source object is, as expected,  strong in $S_\delta$, $S_{CHM}$ and $S_\mu$ but not particularly strong in $S_\ast$. This is due to the dilation implicit encoding of the ``inside'' relationship (identity in this case)
and may not be desired if we want to penalize relationships inside the source.

\paragraph{Closeness.}  Results for the closeness experiment are shown in the second row of Figures~\ref{fig:phi_comparison}, \ref{fig:s_comparison} and \ref{fig:s_midcuts}. As for directional relations, here $S_\ast$ has also attenuated but similar behavior to $S_\delta$, with the notable exception of the source region. In this experiment, even without the extensiveness (as the kernel does not contain the origin),  the effect of the implicit encoding of ``inside'' by the dilation-based score is highly noticeable on $S_\delta$, $S_{CHM}$ and $S_\mu$.

\paragraph{Farness.}  Results for the farness experiment are shown in the third row of Figures~\ref{fig:phi_comparison}, \ref{fig:s_comparison} and \ref{fig:s_midcuts}. As the kernel is too distant from the source for any implicit encoding of ``inside'' to happen, the behavior of $S_\ast$ w.r.t. $S_\delta$ is far more consistent. Interestingly, the most notable difference occurs when $S_\mu$ is close to the source, as it quickly decreases to zero and plateaus there.

\paragraph{Insideness.} Results for the insideness experiment are shown in the fourth row of Figures~\ref{fig:phi_comparison}, \ref{fig:s_comparison} and \ref{fig:s_midcuts}. When ignoring the question of the implicit encoding of the ``inside'' relationship by the dilation-based approach -- by explicitly encoding said relationship -- the most noteworthy observation left from the results is the feathering effect seen in the other experiments.

\subsection{Time and Hyperparameter Comparison}
\begin{figure}[htbp]
    \centering
    \begin{tabular}{|c|c|c|}
        \hline
        Image size & $100 \times 100$ & $200 \times 200$ \\
        \hline
         & 
        \includegraphics[width=0.4\textwidth]{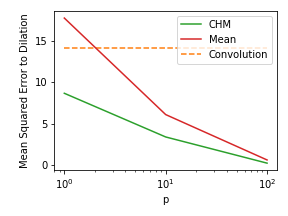} &
        \includegraphics[width=0.4\textwidth]{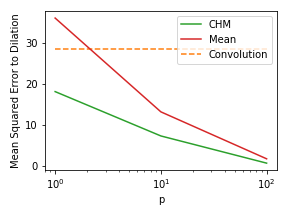} \\
        \hline
    \end{tabular}
    \caption{Results for the hyperparameter comparison experiment, for images of size $100\times 100$ and $200 \times 200$. Dashed lines represent parameter-free methods (dilation and convolutional). Values above $p=10^2$ plateau close to zero. }
    \label{fig:p_comparison_qual}
\end{figure}

\begin{figure}[htbp]
    \centering
    \includegraphics[width=0.6\textwidth]{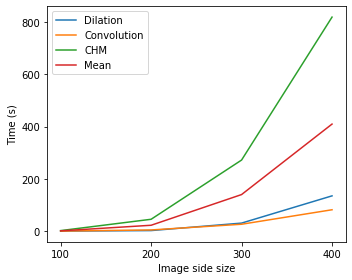}
    \caption{Comparison of the running time of all methods, for different sizes of square images. For the mean approximations, we set $p=100$. }
    \label{fig:time_comparison}
\end{figure}

In order to demonstrate the lighter computational load placed by the convolution operator when compared to the mean approximations, as well as the advantages of its parameter-free implementation, we computed the mean squared error w.r.t. $\Phi_\delta$ for different values of $p$ for each mean approximation, on images of size $100 \times 100$ and of size $200 \times 200$. We also measured the execution time of the computation of $\Phi$ for different resolutions, with a fixed value of $p=100$. Both experiments used the ``to the right of'' relationship as example.
Figure~\ref{fig:p_comparison_qual} highlights the need of the mean based approximations to set an appropriate value of the parameter $p$. 
It can be seen that an improper choice of $p$ may result in sub-optimal results, a problem that the parameter-less convolutional approach does not have.
Figure~\ref{fig:time_comparison} shows that the convolutional approximation vastly outperforms the other approximations in terms of computational time. Additionally, as resolution increases, the gap in computational time increases exponentially. Given the necessity of executing these approximations several times for vast amounts of data when training a neural network, and the modern tendency 
of using input images with high resolutions, these gains in speed are absolutely crucial for completing training in feasible time.

%
%
\section{Conclusion}

We have shown that dilation-based structural constraints can be well enough approximated by a simple, hyper-parameter-free convolution, when used within the context of neural network learning. Additionally, a contextualized comparison with classic methods of the literature has been performed, showcasing their capabilities in this scenario and allowing for a wider variety of choices given computational requirements.

This is the first step towards building a neural network capable of taking advantage of known \textit{a priori} structural relationships, encoded using relational maps, to improve training quality and speed. In the context of neural network training, all assumptions of differentiability made are covered by modern automatic differentiation methods such as those in-built with TensorFlow or PyTorch. From the studied approximations, a working version of a structure-aware training technique for artificial neural networks can be implemented, adding extra terms to the loss function based on the relational scores produced between objects with known expected relationships. 

%
%
%
\bibliographystyle{splncs04}
\bibliography{biblio}

\begin{thebibliography}{1}
\providecommand{\url}[1]{\texttt{#1}}
\providecommand{\urlprefix}{URL }
\providecommand{\doi}[1]{https://doi.org/#1}

\bibitem{Angulo2010-ACIVS-morphconv-chm}
Angulo, J.: Pseudo-morphological image diffusion using the counter-harmonic
  paradigm. In: Blanc-Talon, J., Bone, D., Philips, W., Popescu, D.,
  Scheunders, P. (eds.) Advanced Concepts for Intelligent Vision Systems. pp.
  426--437. Springer Berlin Heidelberg (2010)

\bibitem{IB:FSS-09}
Bloch, I.: Duality vs. {A}djunction for {F}uzzy {M}athematical {M}orphology and
  {G}eneral {F}orm of {F}uzzy {E}rosions and {D}ilations. Fuzzy Sets and
  Systems  \textbf{160},  1858--1867 (2009)

\bibitem{bloch_fuzzy_2005}
Bloch, I.: Fuzzy spatial relationships for image processing and interpretation:
  a review. Image and Vision Computing  \textbf{23}(2),  89--110 (2005)

\bibitem{Dubois1985-IS-aggregation}
Dubois, D., Prade, H.: A {Review} of {Fuzzy} {Set} {Aggregation} {Connectives}.
  Information Sciences  \textbf{36},  85--121 (1985)

\bibitem{Masci2013-MMASIP-morphconv-chm}
Masci, J., Angulo, J., Schmidhuber, J.: A learning framework for morphological
  operators using counter--harmonic mean. In: International Symposium on
  Mathematical Morphology and Its Applications to Signal and Image Processing.
  pp. 329--340. Springer (2013)

\bibitem{Mellouli2019-NNLS-morphconv-chm}
Mellouli, D., Hamdani, T.M., Sanchez-Medina, J.J., Ayed, M.B., Alimi, A.M.:
  Morphological convolutional neural network architecture for digit
  recognition. IEEE Transactions on Neural Networks and Learning Systems
  \textbf{30}(9),  2876--2885 (2019)

\bibitem{murugesan_psi-net:_2019}
Murugesan, B., Sarveswaran, K., Shankaranarayana, S.M., Ram, K., Sivaprakasam,
  M.: {Psi-Net}: {S}hape and boundary aware joint multi-task deep network for
  medical image segmentation. In: 41st Annual International Conference of the
  IEEE Engineering in Medicine and Biology Society (EMBC). pp. 7223--7226
  (2019)

\bibitem{Simantiris2020-STSP}
Simantiris, G., Tziritas, G.: Cardiac {MRI} {Segmentation} with a {Dilated}
  {CNN} {Incorporating} {Domain}-specific {Constraints}. IEEE Journal of
  Selected Topics in Signal Processing  \textbf{14}(6),  1235--1243 (2020)

\end{thebibliography}

\end{document}